\documentclass[10pt,twocolumn,letterpaper]{article}

\usepackage{cvpr}
\usepackage{times}
\usepackage{epsfig}
\usepackage{graphicx}
\usepackage{amsmath}
\usepackage{amssymb}
\usepackage{multirow}
\usepackage{algorithm}
\usepackage{algpseudocode}

\usepackage[breaklinks=true,bookmarks=false]{hyperref}

\cvprfinalcopy 

\def\cvprPaperID{4} 

\def\rot#1{\rotatebox{90}{#1}}

\newcommand{\vect}[1]{\boldsymbol{#1}}

\ifcvprfinal\fi
\begin{document}

\title{Gaussian Process Domain Experts \\for Model Adaptation in Facial Behavior Analysis}

\author{
\begin{tabular}{cccc}
Stefanos Eleftheriadis$^*$ & ~~~~~~~~Ognjen Rudovic$^*$ & ~~~~~~~~Marc P. Deisenroth$^*$ &~~~~~~~~~~~~Maja Pantic$^{*\dagger}$
\end{tabular}\\
$^*$Department of Computing, Imperial College London, UK\\
$^\dagger$EEMCS, University of Twente, The Netherlands\\
{\tt\small \{\href{mailto:s.eleftheriadis@imperial.ac.uk}{s.eleftheriadis}, \href{mailto:orudovic@imperial.ac.uk}{orudovic}, \href{mailto:m.deisenroth@imperial.ac.uk}{m.deisenroth}, \href{mailto:m.pantic@imperial.ac.uk}{m.pantic}\}@imperial.ac.uk}
}

\maketitle

\begin{abstract}
We present a novel approach for supervised domain adaptation that is based upon the probabilistic framework of Gaussian processes (GPs). Specifically, we introduce domain-specific GPs as local experts for facial expression classification from face images. The adaptation of the classifier is facilitated in probabilistic fashion by conditioning the target expert on multiple source experts. Furthermore, in contrast to existing adaptation approaches, we also learn a target expert from available target data solely. Then, a single and confident classifier is obtained by combining the predictions from multiple experts based on their confidence. Learning of the model is efficient and requires no retraining/reweighting of the source classifiers. We evaluate the proposed approach on two publicly available datasets for multi-class (MultiPIE) and multi-label (DISFA) facial expression classification. To this end, we perform adaptation of two contextual factors: `where' (view) and `who' (subject). We show in our experiments that the proposed approach consistently outperforms both source and target classifiers, while using as few as 30 target examples. It also outperforms the state-of-the-art approaches for supervised domain adaptation.
\end{abstract}

\section{Introduction}

Human face is believed to be the most powerful channel for conveying, non-verbally, behavioral traits such as personality, intentions and affect, among others~\cite{ambady1992thin,zeng2009survey}. Facial expressions can be studied at the message level (interpretation in terms of the message conveyed, \eg, emotions), and sign level (analysis of facial muscle movements named action units (AUs)). To this end, the Facial Action Coding System (FACS)~\cite{ekman2002facial} has been used. It is the most comprehensive anatomically-based system for describing facial expressions at both the levels. FACS defines 33 unique AUs, and several categories of head/eye movements.
\looseness-1

\begin{figure}[t]
\centering
\footnotesize
\includegraphics[scale=.48]{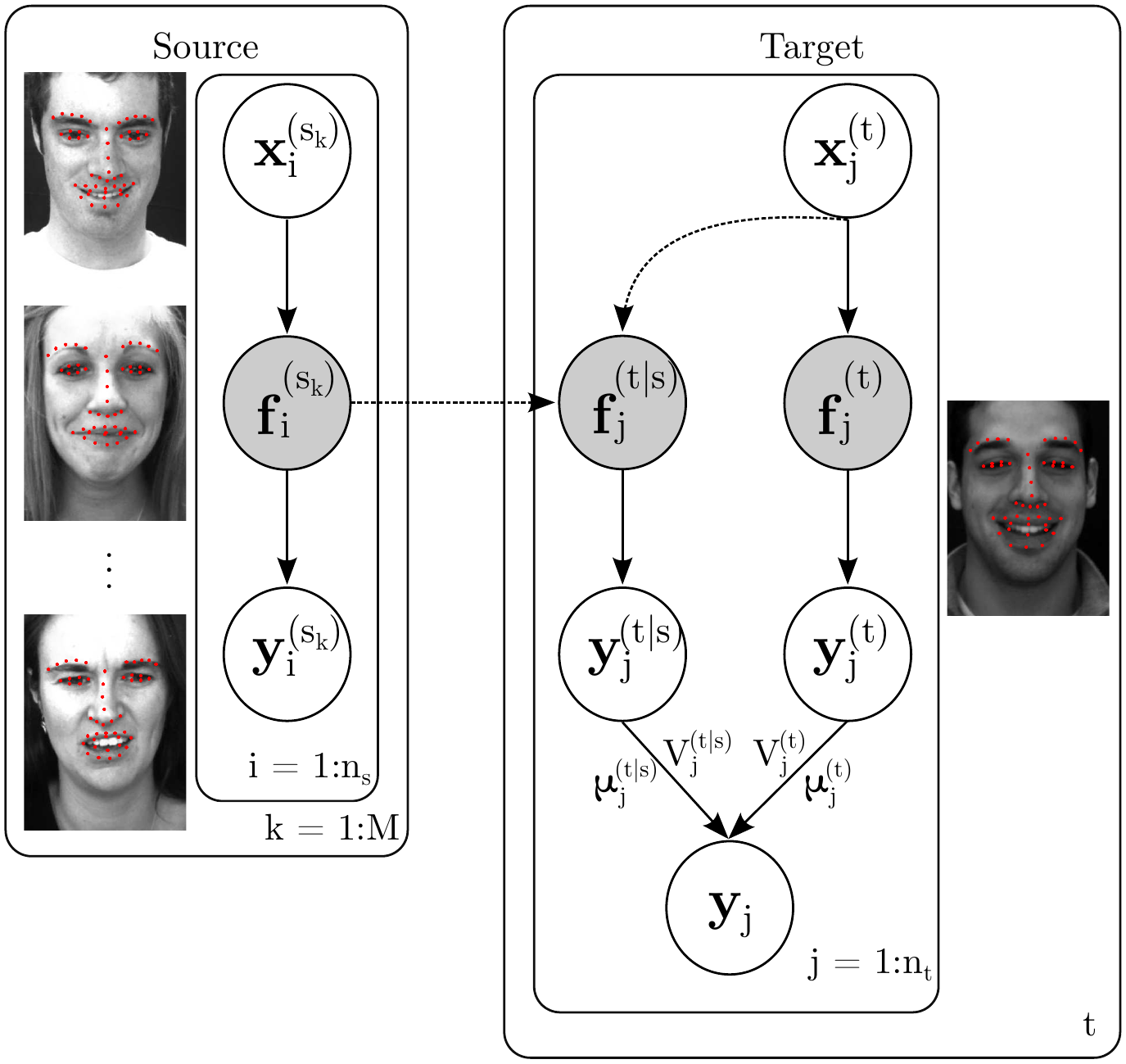}
\caption{{\footnotesize The proposed GPDE model. The learning consists of training the multiple source ($s_k, k=1,\cdots,M$) and the target $(t)$ GP experts (in this case, each subject is treated as an expert), using the available labeled training data pairs $(\vect{x}, \vect{y})$ -- the input features (\eg, facial landmarks) and output labels (\eg, AU activations), respectively. Adaptation (dashed lines) for the target data is performed via conditioning the latent functions, $\vect{f}$, of the target GP on the source experts ($t|s$). During inference, we fuse the predictions from the experts ($ \vect{\mu}^{\{t,(t|s)\}}$) by means of their predictive variance ($ V^{\{t,(t|s)\}}$), with the role of a confidence measure.}}
\label{fig_plate}
\end{figure}

\looseness-1
Due to its practical importance in medicine, marketing and entertainment, automated analysis of facial expressions has received significant research attention over the last two decades. Despite rapid advances in computer vision and machine learning, majority of the models proposed so far for facial expression analysis rely on generic classifiers. These classifiers are expected to generalize well when applied to data recorded within specific contexts, as defined by the W5+ context questions (`who', `where', `how', `what', `when' and `why')~\cite{rudovic2015context}. Nevertheless, due to possible variations in these contextual dimensions, the performance of virtually all existing generic classifiers for facial expression analysis is expected to downgrade largely when applied to previously unseen data~\cite{girard2015how}. This is especially pronounced in the case of unseen subjects (due to variation in their age, gender, expressiveness), changes in pose and illumination, environments, and so on. To circumvent these challenges, two lines of work have been proposed. The first relies on carefull design of 'context-independent' image features and the use of generic classifiers~\cite{zhu2006robust,hu2008study,moore2011local}, while the second attempts adaptation of the target classifiers~\cite{rudovic2015context,chu2013selective,sangineto2014we}. In this work, we employ the latter approach and focus on adaptation of the context questions `where' and `who' in our data.

\looseness-1
Variation in head-pose and illumination (`where') has been addressed by combining illumination invariant features with multi-view learning techniques~\cite{zhu2006robust,hu2008study,moore2011local,rudovic2012coupled,hesse2012multi,eleftheriadis2015discriminative}. On the other hand, the individual differences among subjects (`who') have mainly been tackled by accounting for the subject information at the training stage. Specifically, the original feature set is extended by adding the subject-specific features~\cite{rudovic2015context}, or by building person-specific classifiers~\cite{valstar2011first}. Although these approaches showed improvement over generic classifiers, there is still a number of challenges to address. In particular, the multi-view learning requires a large amount of images in various poses, which is typically not available. On the other hand, for building personalized classifiers, access to an adequate collection of images of the target person is essential. Consequently, existing approaches perform re-weighting previosly learned classifiers to fit the target data (\eg, \cite{chu2013selective}), or training of new models using the additional target data. However, both of these are sub-optimal. Thus, our aim is to find an effective approach to adapt the already trained generic models for facial behavior analysis by using a small number of target data. In the case of the context question `where', this boils down to adapting the frontal classifier to a non-frontal view using only a small number of expressive images from the target view. Similarly, in the case of the subject adaptation (`who'), the model adaptation is performed by using as few annotated images of target subject as needed to gain in the prediction performance (\eg, AU detection). This approach is expected to generalize better than generic classifiers learned from the available source and/or target (training) data.


To address the challenges mentioned above, we use the notion of {\em domain adaptation} to perform two tasks: (i) view and (ii) subject adaptation, for facial expression recognition (FER) and AU detection. In particular, we address the problem of domain adaptation where the distribution of the (facial) features varies across domains (\ie, contexts such as the view or subject), while the output labels (in our case, the emotion or AU activations) remain the same. This is also known as {\em covariate shift}, and the two domains are called {\em source} (\eg, frontal view) and {\em target} (\eg, profile view) domain, respectively. Furthermore, a {\em supervised} setting, where a small number of labeled target examples is available during the adaptation process, is assumed. We build our model upon the probabilistic framework of Gaussian processes (GPs)~\cite{rasmussen2006gaussian}, and generalize the product of expert models~\cite{deisenroth2015distributed,cao2014generalized} to domain adaptation scenario. More specifically, instead of adjusting the classifier parameters between the domains, as in~\cite{chu2013selective,zen2014unsupervised,chen2013learning,miao2012cross,sangineto2014we}, we propose domain-specific GP experts that model the domain specific data.\footnote{The use of GPs for this task is motivated by their good generalization abilities, even when trained with limited amount of data~\cite{rasmussen2006gaussian,eleftheriadis2015discriminative}. This property is crucial for the training of the target expert, since the available data are scarce.} Moreover, instead of minimizing the error between the distributions of the original source and target domain data~\cite{chu2013selective,miao2012cross}, we use Bayesian domain adaptation~\cite{liu2015bayesian} and explain the target data by conditioning on the learned source experts. An advantage of our probabilistic formulation is that during adaptation, we exploit the variance in the predictions when combining the source and target domains~\cite{seeger2003bayesian}. This results in a {\em confident} classifier that minimizes the risk of potential negative transfer (\ie, the adapted model performing worse than the model trained using the adaptation data only).
In contrast to transductive adaptation approaches (\eg, \cite{chu2013selective}) that need to be re-trained completely, adaptation of our model is efficient and requires no re-training of the source model. The model outline is depicted in Fig.~\ref{fig_plate}.  The contributions of this work can be summarized as follows:\looseness-1
\begin{itemize}
\item We present a novel approach for supervised domain adaptation that can, for the first time, perform adaptation to contextual factors `where' (across different views) and `who' (by personalizing the target classifier) during modeling of facial expression data.  
\item To the best of our knowledge, this is the first work in the domain of facial behavior modeling that can simultaneously perform adaption to multiple outputs (\ie, AUs). Existing models in the field that attempt the model adaptation do so for each output independently. 
\item Due to its probabilistic nature, the proposed approach provides the confidence in the predicted labels for the target expressions. This is in contrast to majority of the models that are purely discriminative, and thus, cannot provide a measure of how `reliable' the predictions are.
\item We show in our experiments on view and subject adaptation that the proposed model can generalize better than source and target domains together by using as few as $30$ target samples to perform the adaptation. Furthermore, virtually all existing domain adaptation approaches fail to reach the performance of the target classifiers when more target data become available (negative transfer). Our approach overcomes this due to the newly introduced scheme for combining the source and target experts. 
\end{itemize}

\section{Related Work}

\subsection{Domain Adaptation in Facial Behavior Analysis}
The majority of approaches for domain adaptation in the context of facial behavior analysis focus on building personalized classifiers for the test subjects. For instance,~\cite{miao2012cross} uses the supervised kernel mean matching (KMM) to align the source and target data distributions. This is achieved by re-weighting the source data, which, in combination with the target data, form the input features that are used to train the support vector machine (SVM)~\cite{cortes1995support} classifier for FER. Likewise,~\cite{chu2013selective} uses unsupervised KMM to learn person-specific AU detectors. This is attained by modifying the SVM cost function to account for the KMM between source and target data, adjusting the SVM's hyperplane to the target test data. However, this results in the transductive learning approach, thus, the classifier has to be re-learned for each target subject. In~\cite{chen2013learning}, a two-step learning approach is proposed for person-specific pain recognition and AU detection. First, data of each subject are regarded as different source domains, and are used to train weak Adaboost classifiers. Then, the weak classifiers are weighted based on their classification performance on the available target data. In ~\cite{sangineto2014we,chu2013selective}, the Adaboost classifiers are replaced with the linear SVMs, and then the support vector regression (SVR) is employed to learn the mapping from the feature distribution to the parameters of the SVM classifier.
\looseness-1

Note that, apart from~\cite{chen2013learning}, all the works mentioned above perform in the unsupervised adaptation setting. While this requires less effort in terms of obtaining the labels for the target sub-sample, its underlying assumption is that target data can be well represented as a weighted combination of the source data. However, in real-world data, this assumption can easily be violated, resulting in poor performance of the adapted classifier. In this work, we adopt a supervised approach that needs only a few annotated data from target domain to perform the adaptation. This, in turn, allows us to define both target and source experts, assuring that the performance of the resulting classifier is not constrained by the distribution of the source data, as in unsupervised adaptation approaches. Contrary to transductive learning approaches such as ~\cite{chu2013selective}, our approach requires adaptation of the target expert solely, without the need to re-learn the source experts, resulting in an efficient adaptation process. Moreover, in contrast to our approach, none of the aforementioned works provides a measure of confidence in the predicted labels. Finally, note that the proposed approach and the methods mentioned above differ from those recently proposed for transfer learning~\cite{almaev2015learning}. The goal of the latter is to adapt a classifier learned for, \eg, one AU to another AU, which is different from the adaptation task addressed in this work.
\looseness-1

\subsection{Domain Adaptation}
Domain adaptation is a well studied problem in machine learning (for an extensive survey, see \cite{patel2015visual}). Here we review relevant (semi-)supervised adaptation approaches. For instance, \cite{kulis2011you} learns a transformation that maximizes similarity between data in the source and target domains by enforcing data pairs with the same labels to have high similarity, and pairs with different labels to be dissimilar. Then, a k-NN classifier is used to perform classification of target data.~\cite{hoffman2012discovering} is an extension of this approach to multiple source domains. The input data are assumed to be generated from category-specific local domain mixtures, the mixing weights of which determine the underlying domain of the data, classified using an SVM classifier. Similarly,~\cite{hoffman_ICLR2013} learns a linear asymmetric transformation to maximally align target features to the source domain. This is attained by introducing max-margin constraints that allow the learning of the transformation matrix and SVM classifier jointly.~\cite{donahue2013semi} extends the work in~\cite{hoffman_ICLR2013} by introducing additional constraints to the max-margin formulation. More specifically, unlabeled data from the target domain are used to enforce the classifier to produce similar predictions for similar target-source data. While these methods attempt to directly align the target to source features, several works attempted this through a shared manifold. For instance, ~\cite{duan2012learning} learns a non-linear transformation from both source and target data to a shared latent space, along with the target classifier. Likewise,~\cite{yao2015semi} finds a low-dimensional subspace, which preserves the structure across the domains. The subspace is facilitated by projection functions that are learned jointly with the linear classifier. Again, the structure preservation constraints are used to ensure that similar data across domains are close in the subspace.
\looseness-1

All of the above methods tackle the adaptation problem in a deterministic fashion, thus they do not provide a measure of confidence in the target predictions. By contrast, our approach is fully probabilistic and non-parametric due to the use of GPs. The proposed is related to recent advances in the GP literature~\cite{liu2015bayesian,kandemir2015asymmetric} on domain adaptation. Specifically, in~\cite{liu2015bayesian}, the predictive distribution of a GP trained on the source data is used as a prior for making inference in the target domain. Similarly,~\cite{kandemir2015asymmetric} proposed a two-layer GP that jointly learns separate discriminative functions from the source and target features to the labels. The intermediate layer facilitates the adaptation step, and variational approximation is  employed to integrate out this layer. In contrast to~\cite{liu2015bayesian}, the proposed defines a target specific expert, which is then combined in a principled manner with the source domain experts. The benefit of this is that the resulting classifier is not limited by the distribution of the source data. Also, in contrast to~\cite{kandemir2015asymmetric}, the training of the experts is performed independently, and thus, we need not retrain the source classifier.
\looseness-1


\section{Problem Formulation}\label{prob_def}
We consider a supervised setting for domain adaptation, where we have access to a large collection of labeled {\em source} domain data, and a smaller set of labeled {\em target} domain data. Let $\vect{\mathcal{X}}$ and $\vect{\mathcal{Y}}$ be the input (features) and output (labels) spaces, respectively. We assume that the input space is comprised of the source and target domains, $\mathcal{S}$ and $\mathcal{T}$, respectively, that may differ in feature distribution. Hence, $\vect{X}^{(s)} = \{\vect{x}^{(s)}_{n_s}\}_{n_s=1}^{N_s}$ and $\vect{X}^{(t)} = \{\vect{x}^{(t)}_{n_t}\}_{n_t=1}^{N_t}$, with $\vect{x}^{(s)}_{n_s}, \vect{x}^{(t)}_{n_t}\in \mathbb{R}^{D}$, and $N_t\ll N_s$. In our case, these can be different views or subjects. On the other hand, $\vect{Y}^{(s)}=\{\vect{y}^{(s)}_{n_s}\}_{n_s=1}^{N_s}$ and $\vect{Y}^{(t)}=\{\vect{y}^{(t)}_{n_t}\}_{n_t=1}^{N_t}$ correspond to same labels for both source and target domains. Each vector $\vect{y}^{\{s,t\}}_n$ contains the binary class labels of $C$ classes. We now formulate the regression problem as:
\begin{equation}
 \vect{y}_{n_v}^{(v)} = f^{(v)}(\vect{x}_{n_v}^{(v)}) + \epsilon^{(v)},
\end{equation}
where $\epsilon^{(v)}\sim \mathcal{N}(0, \sigma^2_{v})$ is i.i.d. additive Gaussian noise, and the index $v \in\{s,t\}$ denotes the dependence on each domain. The objective is to infer the latent functions $f^{(v)}$, given the training dataset $\mathcal{D}^{(v)} = \{ \vect{X}^{(v)}, \vect{Y}^{(v)} \}$. By following the framework of GPs~\cite{rasmussen2006gaussian}, we place a prior on the
functions $f^{(v)}$, so that the function values $\vect{f}_{n_v}^{(v)} = f^{(v)}(\vect{x}_{n_v}^{(v)})$ follow a Gaussian distribution $p(\vect{F}^{(v)}|\vect{X}^{(v)}) = \mathcal{N}(\vect{F}^{(v)}|\vect{0}, \vect{K}^{(v)})$. Here, $\vect{F}^{(v)} = \{\vect{f}_{n_v}^{(v)}\}_{n_v=1}^{N_v}$, and $\vect{K}^{(v)} = k^{(v)}(\vect{X}^{(v)},\vect{X}^{(v)})$ is the kernel covariance function, which is assumed to be shared among the label dimensions. In this work, we use the radial basis function (RBF) kernel
\begin{equation}
k(\vect{x},\vect{x}^\prime) = \sigma_f^2\exp\Big(-\frac{1}{2\ell^2}\|\vect{x} - \vect{x}^\prime\|^2\Big),
\end{equation}
where $\{\ell, \sigma_f\}$ are the kernel hyper-parameters. The regression mapping can be fully defined by the set of hyper-parameters $\vect{\theta} = \{\ell, \sigma_f, \sigma_v\}$. Training of the GP consists of finding the hyper-parameters that maximize the log-marginal likelihood
{\small
\begin{align}
\small
\nonumber\log p(\vect{Y}^{(v)}|\vect{X}^{(v)}, \vect{\theta}^{(v)}) =&-\frac{1}{2}\textrm{tr}\left[(\vect{K}^{(v)} + \sigma_v^2\vect{I})^{-1}\vect{Y}^{(v)}{\vect{Y}^{(v)}}^T\right] \\
\label{marg}  &- \frac{C}{2}\log\vert \vect{K}^{(v)} + \sigma_v^2\vect{I} \vert + \textrm{const}.
\end{align}
}
Given a test input $\vect{x}_\ast^{(v)}$ we obtain the GP predictive distribution by conditioning on the training data $\mathcal{D}^{(v)}$ as $p(\vect{f}_\ast^{(v)} |\vect{x}_\ast^{(v)}, \mathcal{D}^{(v)}) = \mathcal{N}(\mu^{(v)}(\vect{x}_\ast^{(v)}), V^{(v)}(\vect{x}_\ast^{(v)}))$ with
\begin{align}
\small
\label{post_mu}\mu^{(v)}(\vect{x}_\ast^{(v)}) &= {\vect{k}_\ast^{(v)}}^T (\vect{K}^{(v)} + \sigma_v^2\vect{I})^{-1}\vect{Y}^{(v)} \\ 
\label{post_s}V^{(v)}(\vect{x}_\ast^{(v)}) &= k_{\ast\ast}^{(v)} - 
{\vect{k}_\ast^{(v)}}^T (\vect{K}^{(v)} + \sigma_v^2\vect{I})^{-1}
\vect{k}_\ast^{(v)},
\end{align}
where {\small $\vect{k}_\ast^{(v)} = k^{(v)}(\vect{X}^{(v)}, \vect{x}_\ast^{(v)})$} and {\small $k_{\ast\ast}^{(v)} = k^{(v)}(\vect{x}_\ast^{(v)}, \vect{x}_\ast^{(v)})$}. For convenience we denote 
$\vect{\mu}_\ast^{(v)} = \mu^{(v)}(\vect{x}_\ast^{(v)})$ and $V^{(v)}_{\ast\ast} = V^{(v)}(\vect{x}_\ast^{(v)})$.
Within the introduced notation, we have the choice to learn either (i) independent functions $f^{(v)}$ or (ii) a universal function $f$ that couples the data from the two domains.
However, neither option allows us to explore the idea of domain adaptation: In the former we learn domain-specific models, while in the latter we simplify the problem by concatenating the data from the two domains.
\looseness-1


\section{Domain Conditioned GPs}
\subsection{GP Adaptation}
A straightforward approach to obtain a model capable of performing inference on data from both domains is to assume the existence of a universal latent function with a single set of hyper-parameters $\vect{\theta}$. To this end, the authors in~\cite{liu2015bayesian} proposed a simple, yet effective, three-step approach for GP adaptation (GPA):
\begin{enumerate}
\item 
Train a GP on the source data with likelihood $p(\vect{Y}^{(s)}|\vect{X}^{(s)}, \vect{\theta})$ to learn the hyper-parameters $\vect{\theta}$. The posterior distribution is the given by Eqs.~(\ref{post_mu}--\ref{post_s}). 
\item Use the obtained posterior distribution of the source data, as a prior for the GP of the target data $p(\vect{Y}^{(t)}|\vect{X}^{(t)}, \mathcal{D}^{(s)}, \vect{\theta})$. 
\item Correct the posterior distribution to account for the target data $\mathcal{D}^{(t)}$ as well.
\end{enumerate}

The prior of the target data in the second step is given by applying Eqs.~(\ref{post_mu}--\ref{post_s}) on $\vect{X}^{(t)}$
\begin{align}
\small
\label{prior_mut}\vect{\mu}^{(t|s)} &= {\vect{K}_{st}^{(s)}}^T (\vect{K}^{(s)} + \sigma_s^2\vect{I})^{-1}\vect{Y}^{(s)} \\ 
\label{prior_st}\vect{V}^{(t|s)} &= \vect{K}_{tt}^{(s)} - 
{\vect{K}_{st}^{(s)}}^T (\vect{K}^{(s)} + \sigma_s^2\vect{I})^{-1}
\vect{K}_{st}^{(s)},
\end{align}
where {\small $\vect{K}_{tt}^{(s)} = k^{(s)}(\vect{X}^{(t)}, \vect{X}^{(t)}), \vect{K}_{st}^{(s)} = k^{(s)}(\vect{X}^{(s)}, \vect{X}^{(t)})$}, and the superscript $t|s$ denotes the conditioning order. Given the above prior and a test input $\vect{x}_\ast^{(t)}$, the correct form of the adapted posterior after observing the target domain data is given by:
{\small
\begin{align}
\small
\label{post_muad}\mu_{ad}^{(s)}(\vect{x}_\ast^{(t)}) &= \vect{\mu}^{(s)}_\ast +  {\vect{V}_\ast^{(t|s)}}^T (\vect{V}^{(t|s)} + \sigma_s^2\vect{I})^{-1}(\vect{Y}^{(t)} - \vect{\mu}^{(t|s)}) \\ 
\label{post_sad}V_{ad}^{(s)}(\vect{x}_\ast^{(t)}) &= V_{\ast\ast}^{(s)} - 
{\vect{V}_\ast^{(t|s)}}^T (\vect{V}^{(t|s)} + \sigma_s^2\vect{I})^{-1}
\vect{V}_\ast^{(t|s)},
\end{align}
}
with {\small $\vect{V}_\ast^{(t|s)} = k^{(s)}(\vect{X}^{(t)}, \vect{x}_\ast^{(t)}) - 
{k^{(s)}(\vect{X}^{(s)}, \vect{X}^{(t)})}^T (\vect{K}^{(s)} + \sigma_s^2\vect{I})^{-1} k^{(s)}(\vect{X}^{(s)}, \vect{x}_\ast^{(t)})$}.

Eqs.~(\ref{post_muad}--\ref{post_sad}) shows that final prediction in the GPA is the combination of the original prediction based on the source data only, plus a correction term. The latter shifts the mean toward the distribution of the target data and improves the model's confidence by reducing the predictive variance.
Note that we originally constrained the model to learn a single latent function $f$ for  both conditional distributions $p(\vect{Y}^{(v)}|\vect{X}^{(v)})$ to derive the posterior for the GPA. However, this constraint  implicitly assumes that the marginal distributions of the data $p(\vect{X}^{(v)})$ are similar. This assumption violates the general idea of domain adaptation, where by definition, the marginals may have significantly different attributes (\eg, input features from different observation views). In such cases, GPA could perform worse than an independent GP trained solely on the target data $\mathcal{D}^{(t)}$. One possible way to address this issue is to retrain the $\log p(\vect{Y}^{(t)}|\vect{X}^{(t)}, \mathcal{D}^{(s)}, \vect{\theta})$ of the GPA \wrt $\vect{\theta}$~\cite{liu2015bayesian}. This option will compensate for the differences in the distributions by readjusting the hyper-parameters. However, it comes with the price of retraining of the model. Furthermore, it does not allow for modeling domain-specific attributes since the predictions are still determined mainly from the source distribution.
\looseness-1

\subsection{GP Domain Experts}

\noindent\textbf{Product of GP Experts}.
In the proposed approach, we assume that each expert is a GP that operates only on a subset of data, \ie, $\mathcal{D}^{(s)}, \mathcal{D}^{(t)}$. Hence, we can follow the methodology presented in Sec.~\ref{prob_def} in order to train domain-specific GPs and learn different latent functions, \ie, hyper-parameters $\vect{\theta}^{(v)}$. Within the current formulation we treat the source domain as a combination of multiple source datasets (\eg, subject-specific datasets) $\mathcal{D}^{(s)} = \{\mathcal{D}^{(s_1)},\dotsc, \mathcal{D}^{(s_M)} \}$, where $M$ is the total number of source domains (datasets).
\looseness-1

\noindent\textbf{Training}.
Given the above mentioned data split and assuming conditional independence, the marginal likelihood can be approximated by
\begin{align}
\nonumber p(&\vect{Y}^{\{s,t\}},| \vect{X}^{\{s,t\}}, \vect{\theta}^{\{s,t\}}) = \\ 
\label{marg_cond}&p(\vect{Y}^{(t)} | \vect{X}^{(t)}, \vect{\theta}^{(t)}) \prod_{k = 1}^M p_k(\vect{Y}^{(s_k)} | \vect{X}^{(s_k)}, \vect{\theta}^{(s)}).
\end{align}
Note that we share the set of hyper-parameters $\vect{\theta}^{(s)}$ across all the source domains. The intuition behind this is that in each source domain we may observe different label distribution $p(\vect{Y}^{(s_k)})$, yet after exploiting all the available datasets we can model the  overall distribution $p(\vect{Y}^{(s)})$ with a single set of hyper-parameters $\vect{\theta}^{(s)}$. However, this does not guarantee that we are also able to explain the target label distribution $p(\vect{Y}^{(t)})$ with the same hyper-parameters. Thus, we also search for $\vect{\theta}^{(t)}$ for modeling the domain-specific attributes. Similarly to Sec.~\ref{prob_def} learning of the hyper-parameters is performed by maximizing
\begin{align}
\nonumber \log p(\vect{Y}^{\{s,t\}},| &\vect{X}^{\{s,t\}}, \vect{\theta}^{\{s,t\}}) = 
\log p(\vect{Y}^{(t)} | \vect{X}^{(t)}, \vect{\theta}^{(t)})\\ 
\label{logmarg_cond}&+ \sum_{k = 1}^M \log p_k(\vect{Y}^{(s_k)} | \vect{X}^{(s_k)}, \vect{\theta}^{(s)}),
\end{align}
where each log-marginal is computed according to Eq.~(\ref{marg}). The above factorization, apart from facilitating learning of the domain experts, allows for efficient GP training even with larger datasets, as shown in~\cite{deisenroth2015distributed}. Note that the source experts can be learned independently from the target, which allows our model to generalize to unseen target domains without retraining.
\looseness-1

\noindent\textbf{Predictions}.
Once we have trained the GPDE, we need to combine the predictions from each expert to form an overall prediction. To this end, we follow the approach presented in~\cite{cao2014generalized}, where we further readjust the predictions from the source experts using the trick of GPA. Hence, the predictive distribution is given by
\begin{align}
\nonumber p(\vect{f}_\ast^{(t)}|\vect{x}_\ast^{(t)},\mathcal{D}) = &\prod_{k = 1}^M p_k^{\beta_{s_k}}(\vect{f}_\ast^{(t)}|\vect{x}_\ast^{(t)},\mathcal{D}^{(s_k)},\mathcal{D}^{(t)},\vect{\theta}^{(s)}) \cdot\\ 
\label{prod_exp}&p^{\beta_t}(\vect{f}_\ast^{(t)}|\vect{x}_\ast^{(t)},\mathcal{D}^{(t)}, \vect{\theta}^{(t)}),
\end{align}
where $\beta_{s_k}, \beta_t$ control the contribution of each expert. In this work we equally weight the experts and normalize them such that $\beta_t + \sum\beta_{s_k} = 1$, as suggested in~\cite{deisenroth2015distributed}. The predictive mean and variance are given by
{\small
\begin{align}
\small
\label{post_mugpde}&\vect{\mu}_\ast^{\textrm{gpde}} = V_\ast^{\textrm{gpde}}\left[\beta_t {V_\ast^{(t)}}^{-1} \vect{\mu}^{(t)}_\ast + {\sum\nolimits}_k  \beta_{s_k} {V_{ad}^{(s_k)}}^{-1} \vect{\mu}^{(s_k)}_{ad}\right] \\ 
\label{post_sgpde}&V_\ast^{\textrm{gpde}} = \left[ \beta_t {V_\ast^{(t)}}^{-1} + {\sum\nolimits}_k  \beta_{s_k} {V_{ad}^{(s_k)}}^{-1}\right]^{-1}.
\end{align}
}
\looseness-1

At this point the contribution of the GPDE becomes clear: Eq.~(\ref{post_mugpde}) shows that the overall mean is the sum of the predictions from each expert, weighted by their precision (inverse variance). Hence, the solution of the GPDE will favor the predictions of more confident experts. On the other hand, if the quality of a domain expert is poor (noisy predictions with large variance), GPDE will weaken its contribution to the overall prediction. Algorithm~\ref{alg} summarizes the GPDE adaptation procedure.
\looseness-1
\begin{algorithm}[t]
\small
\caption{Domain adaptation with GPDE}
\label{alg}
\begin{algorithmic}
\\\hrulefill\\\quad Inputs: $\mathcal{D}^{(s)}=\{\vect{X}^{(s)}, \vect{Y}^{(s)}\}, \mathcal{D}^{(t)}=\{\vect{X}^{(t)}, \vect{Y}^{(t)}\}$
\State \textbf{Training:}
\\\quad Learn the hyper-parameters $\vect{\theta}^{\{s,t\}}$ by maximizing Eq.~(\ref{logmarg_cond}).
\State \textbf{Adaptation:}
\\\quad Adapt the posterior from the source experts via Eq.~(\ref{post_muad}--\ref{post_sad}).
\State \textbf{Predictions of Experts:}
\\\quad Combine the prediction from each GP domain expert via \\\quad Eq.~(\ref{post_mugpde}--\ref{post_sgpde}).
\\\quad Output: $\vect{y}_\ast = sign(\vect{\mu}_\ast^{gpde})$.
\end{algorithmic}
\end{algorithm}


\section{Experiments}

We evaluate the proposed model on acted and spontaneous facial expressions from two publicly available datasets: MultiPIE~\cite{gross2010multi} and Denver Intensity of Spontaneous Facial Actions (DISFA)~\cite{mavadati2013disfa}. 
Specifically, MultiPIE contains images of 373 subjects depicting acted facial expressions of Neutral (NE), Disgust (DI), Surprise (SU), Smile (SM), Scream (SC) and Squint (SQ), captured at various pan angles. In our experiments, we used images from $0^\circ$, $-15^\circ$ and $-30^\circ$. DISFA is widely used in the AU-related literature, due to the large amount of (subjects and AUs) annotated images. It contains video recordings of 27 subjects while watching YouTube videos. Each frame is coded in terms of the AU intensity on a six-point ordinal scale. For our experiments we selected the six most frequently occurring AUs, \ie, AUs {\small(4, 6, 9, 12, 25, 26)}, while we treated each AU with intensity larger than zero as active.
\looseness-1

\noindent\textbf{Features:} We use a set of geometric features derived from the facial landmark locations. DISFA dataset comes with frame-by-frame annotations of 66 facial points, while the same annotated points for MultiPIE were obtained from~\cite{sagonas}. We discarded the contour landmarks, leading to the set of 49 facial points. These were then registered to a reference face (average face per view for MultiPIE, and average face for DISFA) using an affine transform. In order to further remove potential noise and artifacts, the aligned landmark points were post-processed via PCA, retaining 99\% of the energy, which resulted in 30D feature vectors. 
\looseness-1

\begin{table*}[th]
\setlength{\tabcolsep}{4.5pt}  
\footnotesize
\centering
\caption{Average classification rate across 5-folds on MultiPIE. The adaptation is performed from $0^\circ \rightarrow -15^\circ$ and $0^\circ \rightarrow -30^\circ$, with increasing cardinality of labeled target domain data ($10-1200$).}
\begin{tabular}{l||c|c|c|c|c|c|c|c||c|c|c|c|c|c|c|c}
\hline
\hline
\multirow{2}{*}{Method} & \multicolumn{8}{c||}{$0^\circ\rightarrow-15^\circ$} & \multicolumn{8}{c}{$0^\circ\rightarrow-30^\circ$}\\\cline{2-17}
 &10 &30    & 50    & 100  &200 &300 &600 &1200  &10 &30    & 50    & 100  &200 &300 &600 &1200  \\ \hline\hline
GP$_{source}$	&\multicolumn{8}{c||}{81.65}	&\multicolumn{8}{c}{76.94}\\ \hline
GP$_{target}$	&55.85	&81.19	&84.59	&89.61	&90.66	&91.31	&91.57	&97.26  &51.99	&76.09	&81.97	&86.48 &88.57	&89.75	&\textbf{92.16}	&\textbf{98.43}\\ \hline
GPA~\cite{liu2015bayesian}	&82.36	&84.00	&85.37	&88.63 &90.20	&91.51	&93.79	&96.15	&77.73	&79.82	&81.65	&85.43 &87.79	&87.72	&89.29	&93.01\\ \hline
ATL-DGP~\cite{kandemir2015asymmetric}	&\textbf{83.32}	 &86.34   &85.22   &85.62   &85.16   &86.42   &86.53     &87.80	&\textbf{79.82}	&\textbf{82.93}		&83.36   &85.53   &82.08   &84.32   &80.03   &83.04\\ \hline
GPDE	&82.95	&\textbf{86.35}	&\textbf{87.52}	&\textbf{92.10}	&\textbf{93.73}	&\textbf{94.64}	&\textbf{95.36}	&\textbf{97.84} &78.71	&82.17	&\textbf{84.65}	&\textbf{87.85} &\textbf{88.83}	&\textbf{90.01}	&91.38	&96.86\\ \hline\hline
\end{tabular}
\label{tab_views}
\end{table*}

\noindent\textbf{Evaluation procedure.} We evaluate GPDE on both multi-class (FER on MultiPIE) and multi-label (multiple AU detection on DISFA) scenarios. We also assess the adaptation capacity of the model with a single (view adaptation) and multiple (subject adaptation) source domains. For the task of FER, the frontal view, \ie, $0^\circ$, served as a single source domain, and inference was performed via adaptation to the target domains $-15^\circ$ and $-30^\circ$. For the AU detection task, the various subjects from the train data were used as multiple source domains, and adaptation was performed each time on the tested subject. To evaluate the model's adaptation ability we strictly follow a training protocol, where for each experiment we vary the cardinality of the training target data (we always use all the available source domain data). For MultiPIE, we first split the data in 5-folds (4 training, 1 testing) and then, we keep increasing the cardinality as: $N_t = 10, 30, 50, 100, 200, 300, 600, 1200$. For DISFA we partition the data in 3-folds (20 training source subjects at a time). From the test subject's sequence the first 500 frames were used as target training data (with increasing cardinality $N_t = 10, 30, 50, 100, 200, 500$), while inference was performed on the rest frames of the sequence. This is in order to avoid the target model overfitting the temporally neighboring examples of test subject. For the FER experiments, we employ the classification ratio (CR) as the evaluation measure, while for the AU detection, due to the imbalance in the data, we report the F1 score and the area under the ROC curve (AUC).
\looseness-1


\noindent\textbf{Models compared.} We compare the proposed GPDE with the two generic models GP$_{source}$ and GP$_{target}$. The former is trained solely on the source data, while the latter on the target data used for the adaptation. Furthermore, we compare to the state-of-the-art models for supervised domain adaptation, \ie, the GPA~\cite{liu2015bayesian} and the asymmetric transfer learning with deep GP (ATL-DGP)~\cite{kandemir2015asymmetric}. The GPA is an instance of the proposed GPDE, with only a source domain expert (no target) and predictions given by Eq.~(\ref{post_muad}--\ref{post_sad}). ATL-DGP\footnote{The provided code for ATL-DGP is not capable of multi-label classification, since it treats the labels only in a 1-of-K encoding. Thus, it cannot be evaluated on the multiple AU detection task.} employs an intermediate GP to combine the predictions of GP$_{source}$ and GP$_{target}$. In the multi-source experiment we also compare to GPDE$_{ss}$, which is the instance of GPDE with all the subjects treated as a single source domain. Note that We do not include comparisons with the deterministic approaches (\eg,~\cite{hoffman_ICLR2013,kulis2011you}), as it has been shown in~\cite{kandemir2015asymmetric} that ATL-DGP outperforms these methods.
\looseness-1

\subsection{View adaptation from a single source}
\begin{figure}[t]
\centering
\footnotesize
\includegraphics[scale=.8]{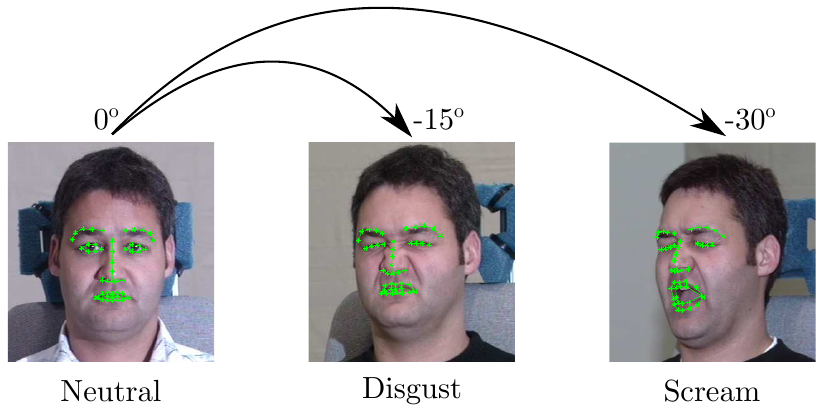}
\caption{{\footnotesize View adaptation for FER on MultiPIE dataset.}}
\label{fig_db}
\end{figure}

In this experiment, we demonstrate the effectiveness of the proposed approach when the distributions between source and target domain ($0^\circ \rightarrow -15^\circ$ and $0^\circ \rightarrow -30^\circ$) differ in an increasing non-linear manner.
For this purpose we evaluate all considered algorithms in terms of their ability to perform accurate FER as we move away from the frontal pose. Example images for the specified task can be seen in Fig.~\ref{fig_db}. Table~\ref{tab_views} summarizes the results. The generic classifier GP$_{source}$ exhibits the lowest performance, due to the fact that it has only been  trained on source domain images. It is important to note the drop in the classification rate ($\approx 5\%$) when the target domain changes from $-15^\circ$ to $-30^\circ$. This indicates the inefficiency of a generic classifier to deal with data of different characteristics. On the other hand, the GP$_{target}$ when trained with as few as $30$ data points achieves similar performance to the GP$_{source}$ since it benefits from modeling domain-specific attributes. A further increase of the cardinality of the target training data results in a significant improve in the classification rate. A similar trend can be observed in the performance of the adaptation methods, where the inclusion of $10$ labeled data points from the target domain is adequate to shift the learned source classifier towards the distribution of the target data. 
\looseness-1

The GPA uses the extra data to condition on the generic classifier GP$_{source}$ and increase its prediction performance. ATL-DGP on the other hand facilitates a joint learning scheme where GP$_{source}$ and GP$_{target}$ are fused together, via conditioning, in a deep architecture. The advantage of the latter is evidenced from the highest achieved accuracy, \ie, $83.32\%$ for $N_t = 10$. However, the joint training scheme of ATL-DGP limits its adaptation ability, due to the high effect of the source prior. Consequently, its performance saturates. A further disadvantage of ATL-DGP's joint learning is that it requires retraining every time the target distribution changes. Finally, the proposed GPDE, uses the notion of experts to unify 
GP$_{source}$ and GP$_{target}$ into a single classifier. To achieve so, GPDE measures the confidence of the predictions from each expert (by means of predictive variance), in contrast to GPA (uses source expert only) and ATL-DGP (uses an uninformative prior). This property of GPDE is more pronounced in the adaptation  $0^\circ \rightarrow -30^\circ$ with $N_t > 300$, where GP$_{target}$ achieves the highest classification ratio. GPDE performs similarly to the target expert while, GPA and ATL-DGP underestimate the prediction capacity of the target-specific classifier, and thus, attain lower results.\looseness-1

\begin{figure*}[th]
\centering
\footnotesize
\begin{tabular}{cccc}
\includegraphics[scale=.33]{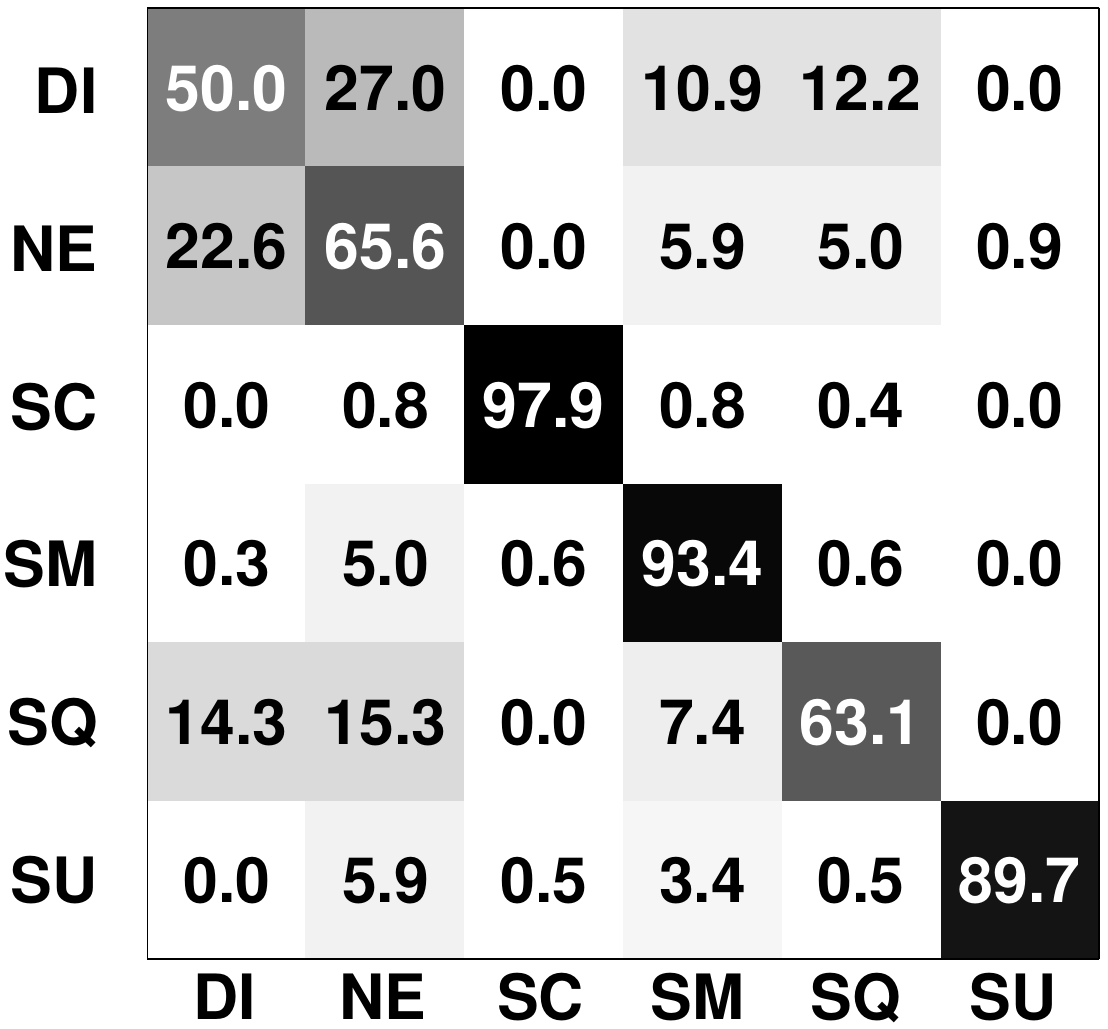} &
\includegraphics[scale=.33]{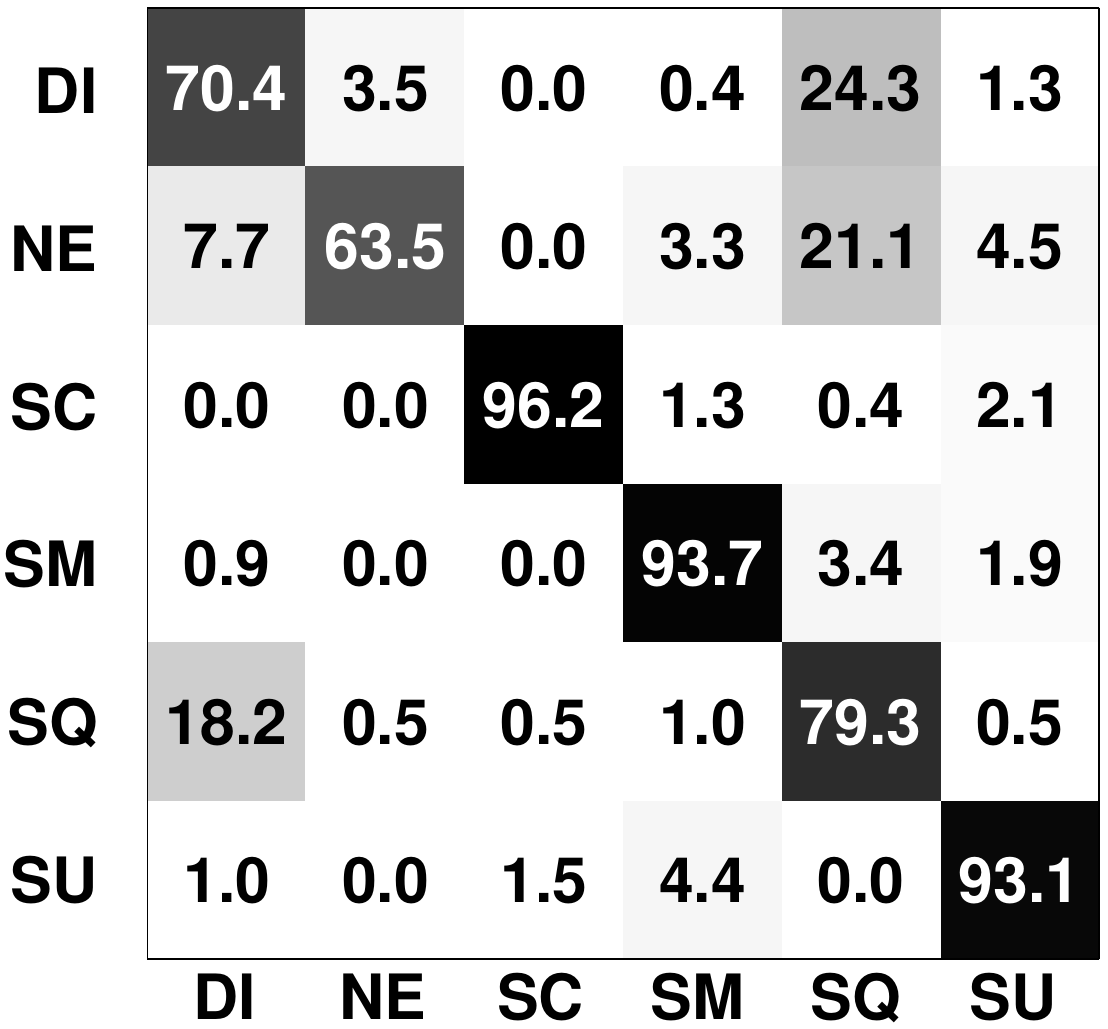} &
\includegraphics[scale=.33]{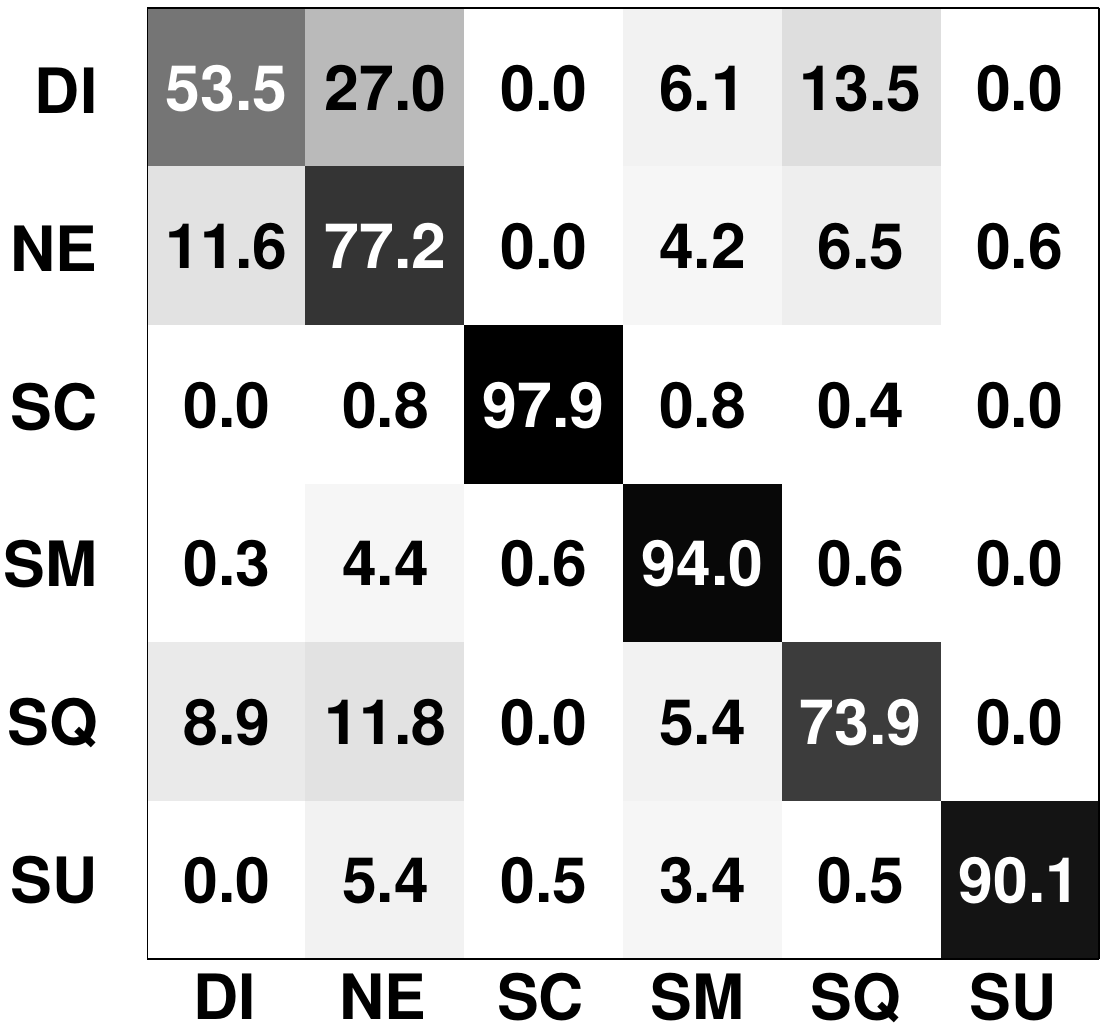} &
\includegraphics[scale=.33]{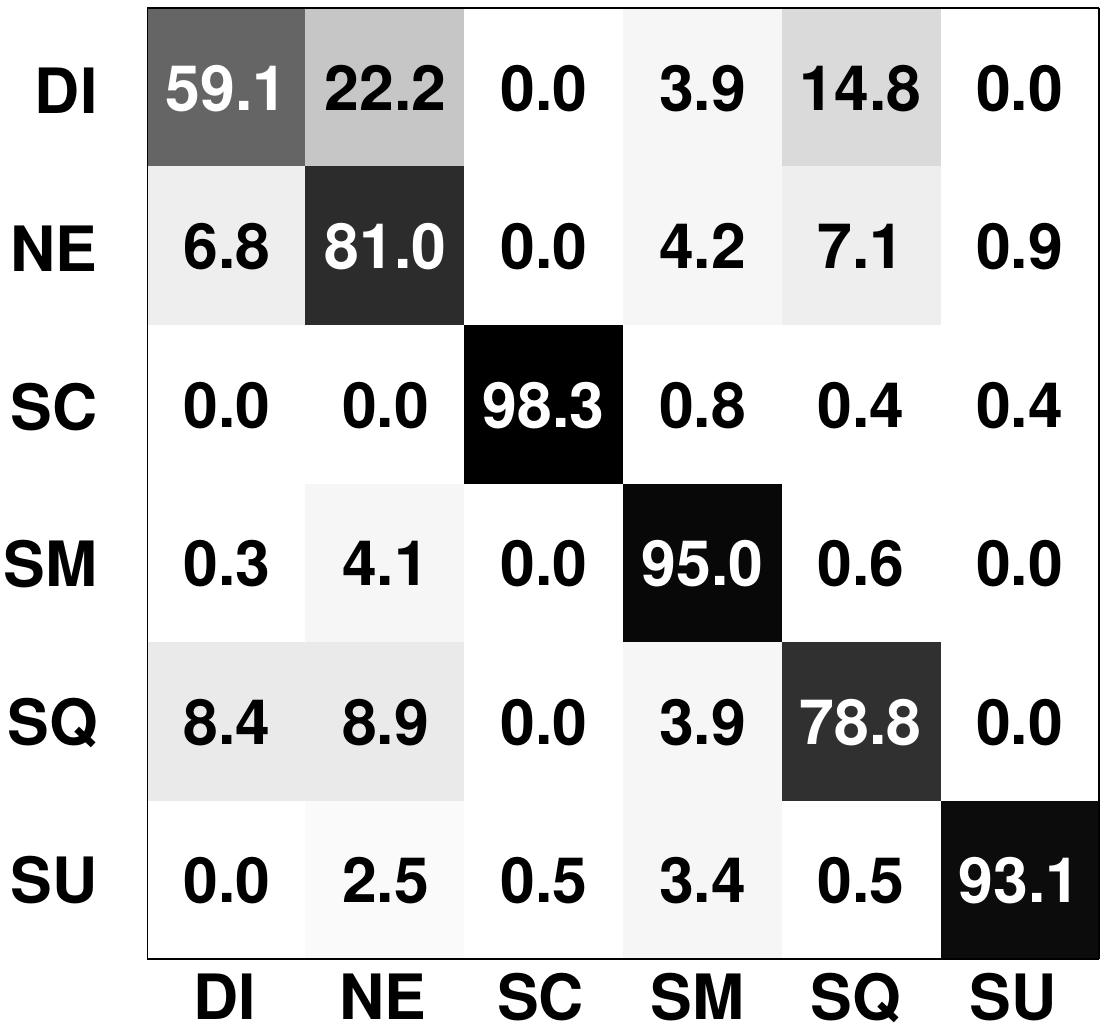}\\
GP$_{source}$ & GP$_{target}$ & GPA & GPDE
\end{tabular}
\caption{{\footnotesize Confusion matrices averaged across the folds when using 50 target training data for $0^\circ \rightarrow -30^\circ$ adaptation.}}
\label{fig_cm}
\end{figure*}

A better insight into the performance of the considered methods can be obtained from the confusion matrices in Fig.~\ref{fig_cm}. The reported results are for $0^\circ \rightarrow -30^\circ$ adaptation with $N_t = 50$ (at which point the GP$_{target}$ starts outperforming GP$_{source}$). The proposed GPDE takes advantage of the target-specific expert and significantly reduces the confusion between the subtle expressions of Disgust and Squint with the Neutral face.
\looseness-1

\begin{table*}[ht]
\setlength{\tabcolsep}{4.5pt}  
\footnotesize
\centering
\caption{Average results of 6 jointly predicted AUs with subject adaptation on DISFA.}
\begin{tabular}{l|c|c|c|c|c|c||c|c|c|c|c|c}
\hline
\hline
\multirow{2}{*}{Method} & \multicolumn{6}{c||}{Average F1 score} & \multicolumn{6}{c}{Average AUC}\\\cline{2-13}
 &10 &30    & 50    & 100   &200	&500	& 10   & 30   & 50    & 100 &200	&500 \\ \hline\hline
GP$_{source}$	& \multicolumn{6}{c||}{53.17} & \multicolumn{6}{c}{74.77}\\ \hline
GP$_{target}$	&52.16	&53.74	&54.36	&55.24 &55.60 &55.06	&70.61	&73.00	&73.84	&74.94 &75.32 &74.59\\ \hline
GPA~\cite{liu2015bayesian}	&56.54	&57.42	&57.85	&57.87	&58.22 &58.39 &76.45	&77.64	&78.14	&78.52 &79.07 &79.38\\ \hline
GPDE$_{ss}$	&56.27	&57.74	&58.49	&58.76 &59.12 &58.88	&75.04	&77.83	&78.72	&79.23 &79.67 &79.08\\ \hline
GPDE	&\textbf{58.66}	&\textbf{60.04}	&\textbf{60.56}	&\textbf{60.18} &\textbf{60.48} &\textbf{60.17}	&\textbf{78.25}	&\textbf{80.15}	&\textbf{80.59}	&\textbf{80.20} &\textbf{80.27} &\textbf{79.86}\\ \hline\hline
\end{tabular}
\label{tab_aus_avg}
\end{table*}

\subsection{Subject adaptation from multiple sources}
In this section, we evaluate the models in a multi-label classification scenario, where the adaptation is performed from multiple source domains. This is a more challenging setting, since the dataset is comprised of naturalistic facial expressions, and the recorded subjects are experiencing the affect in different ways and levels. 
The difficulty of the task can be seen in Table~\ref{tab_aus_avg}, where the subject-specific classifier GP$_{target}$, trained with $30$ labeled data points, achieves a higher F1 score than the generic classifier  GP$_{source}$, which is trained on 20 subjects. The adaptation attained by GPA and GPDE$_{ss}$ (the single source instance of GPDE) results in an improved average score compared to the subject specific GP$_{target}$. At this point note that GPA and GPDE$_{ss}$ perform similarly. The reason for this is that by treating all training subjects as a single source domain, GPDE$_{ss}$ smooths out the individual differences of the training subjects by treating them as data from a  single, \emph{broader}, source domain. Thus, the contribution of the target domain expert is diminished, as the variations of the target data can be explained, on average, by the source domain. On the contrary, the proposed GPDE with the adaptation from multiple sources (one per training subject) not only attains the best average F1 scores, but also achieves a more robust performance as evidenced from the higher AUC. Finally, note that with $N_t = 10$ GPDE performs better than the target specific classifier with $N_t = 500$. Note also that the proposed GPDE reaches the full (and the highest of all) performance with only 30 samples from the target domain. This is an important result, since obtaining the AU annotations (6 in this experiment) is expensive and time consuming.
\looseness-1

Table~\ref{tab_aus} reports the detailed results (F1 score) per AU for the case of $N_t = 50$. The proposed GPDE attains a significant improvement (more than 5\%) in AU4,6,25 compared to its counterparts, while it only suffers a loss from GPA on AU26. Moreover, the ROC curves in Fig.~\ref{fig_auc} show that GPDE exhibits a more robust performance not only on AUs with more pronounced improvement (\ie, AU6), but also on AUs with similar F1 score to GPA (\ie, AU12). The latter indicates that the proposed GPDE is a more robust model.
\looseness-1

\begin{table}[th]
\setlength{\tabcolsep}{4.5pt}  
\footnotesize
\centering
\caption{F1 score for joint AU detection on DISFA. Subject adaptation with $N_t = 50$.}
\label{my-label}
\begin{tabular}{l|c|c|c|c|c|c|c}
\hline
\hline
Method & AU$4$ & AU$6$ & AU$9$ & AU$12$ & AU$25$ & AU$26$ & Avg.\\ \hline\hline
GP$_{source}$	&51.93	&42.34	&41.06	&58.89	&78.84	&57.98	&53.17\\ \hline
GP$_{target}$	&59.85	&48.54	&46.79	&53.23	&63.14	&54.61	&54.36\\ \hline
GPA~\cite{liu2015bayesian}	&56.75	&47.97	&43.88	&\textbf{60.33}	&78.35	&\textbf{59.82} & 57.85\\ \hline
GPDE$_{ss}$	&60.20	&50.90	&\textbf{47.67}	&59.17	&73.82	&59.17	&58.49\\ \hline
GPDE	&\textbf{65.59}	&\textbf{53.62}	&47.10	&60.02	&\textbf{79.96}	&57.08	&\textbf{60.56}\\ \hline\hline
\end{tabular}
\label{tab_aus}
\end{table}

\begin{figure}[th]
\centering
\footnotesize
\setlength{\tabcolsep}{1.5pt}
\begin{tabular}{cc}
\includegraphics[scale=.27]{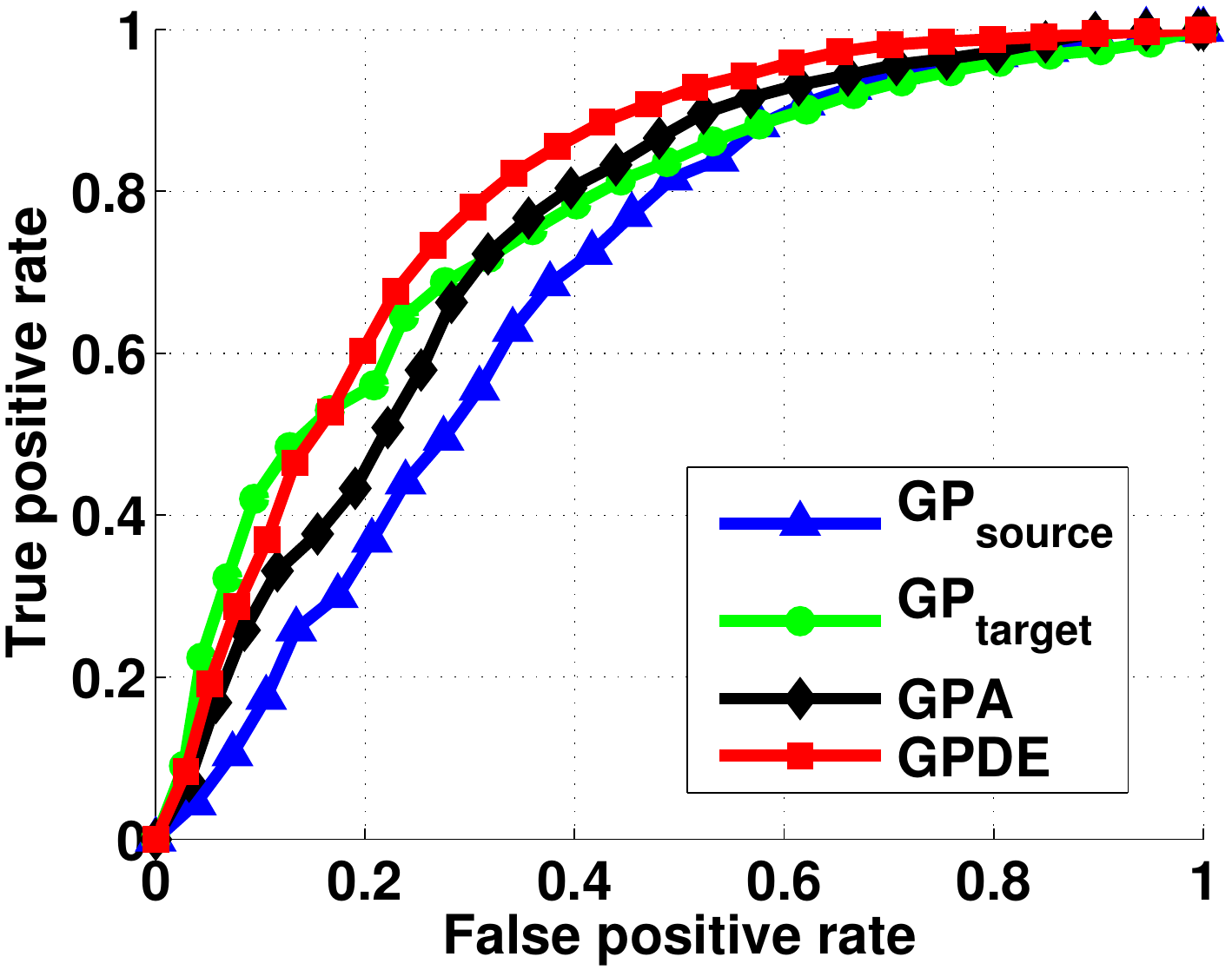} &
\includegraphics[scale=.27]{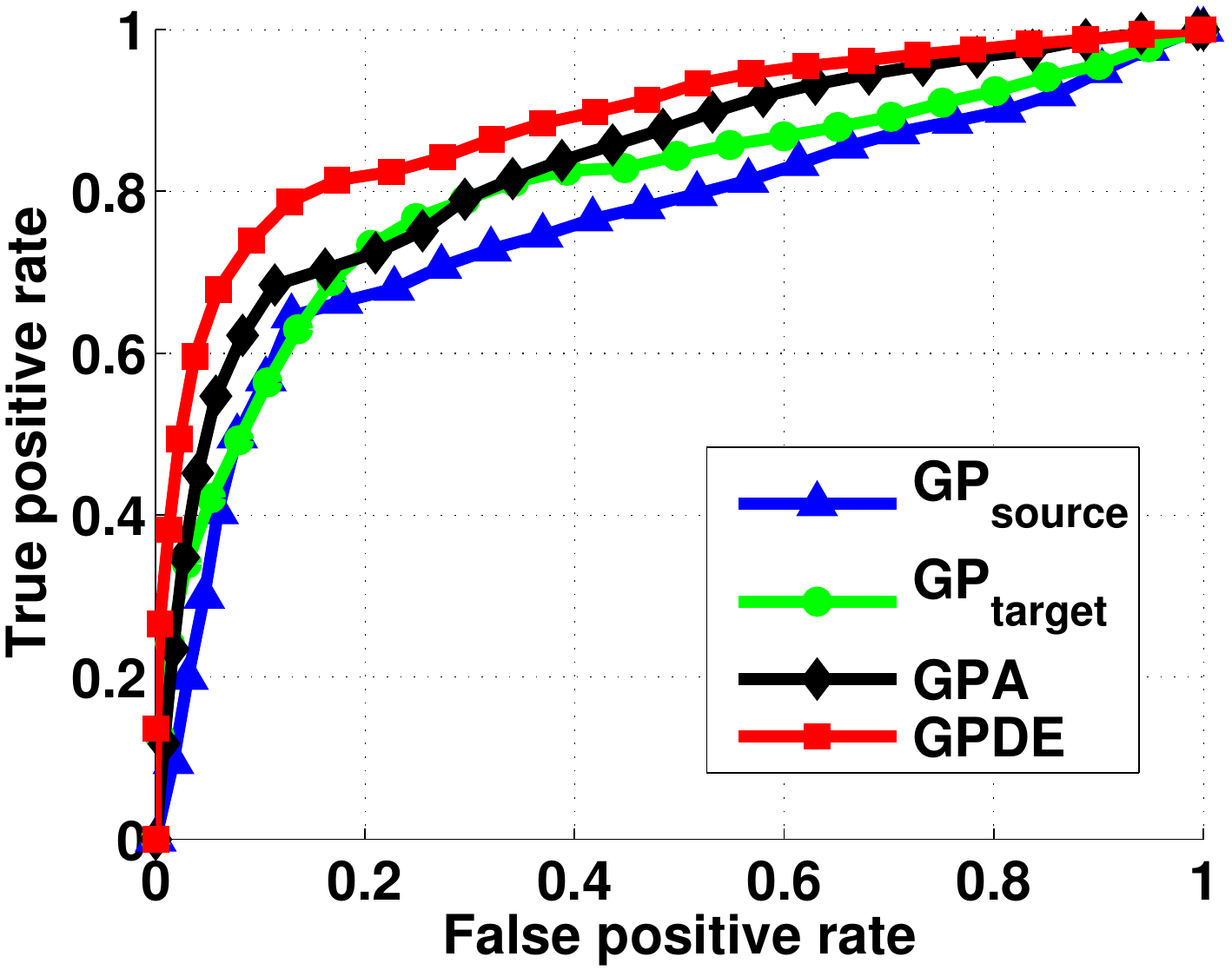}\\
AU$6$ & AU$12$
\end{tabular}
\caption{{\footnotesize Average ROC curves for AU$6$ (left) and AU$12$ (right). Subject adaptation with $N_t = 50$.}}
\label{fig_auc}
\end{figure}

\begin{figure}[th]
\centering
\footnotesize
\setlength{\tabcolsep}{1.5pt}
\begin{tabular}{lcc}
\rot{\rlap{\qquad \quad test subj.\# 21 }} &\includegraphics[scale=.27]{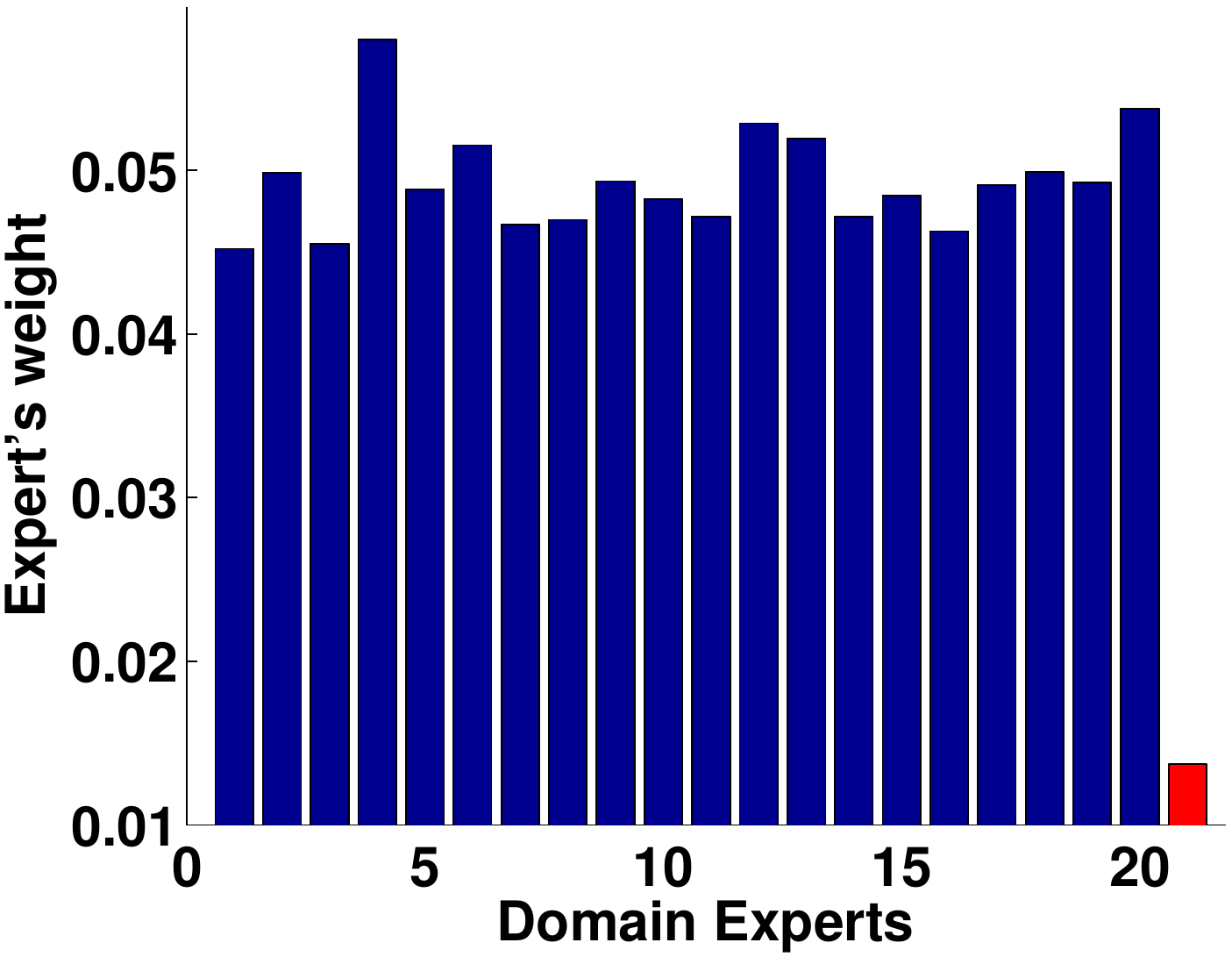} &
\includegraphics[scale=.27]{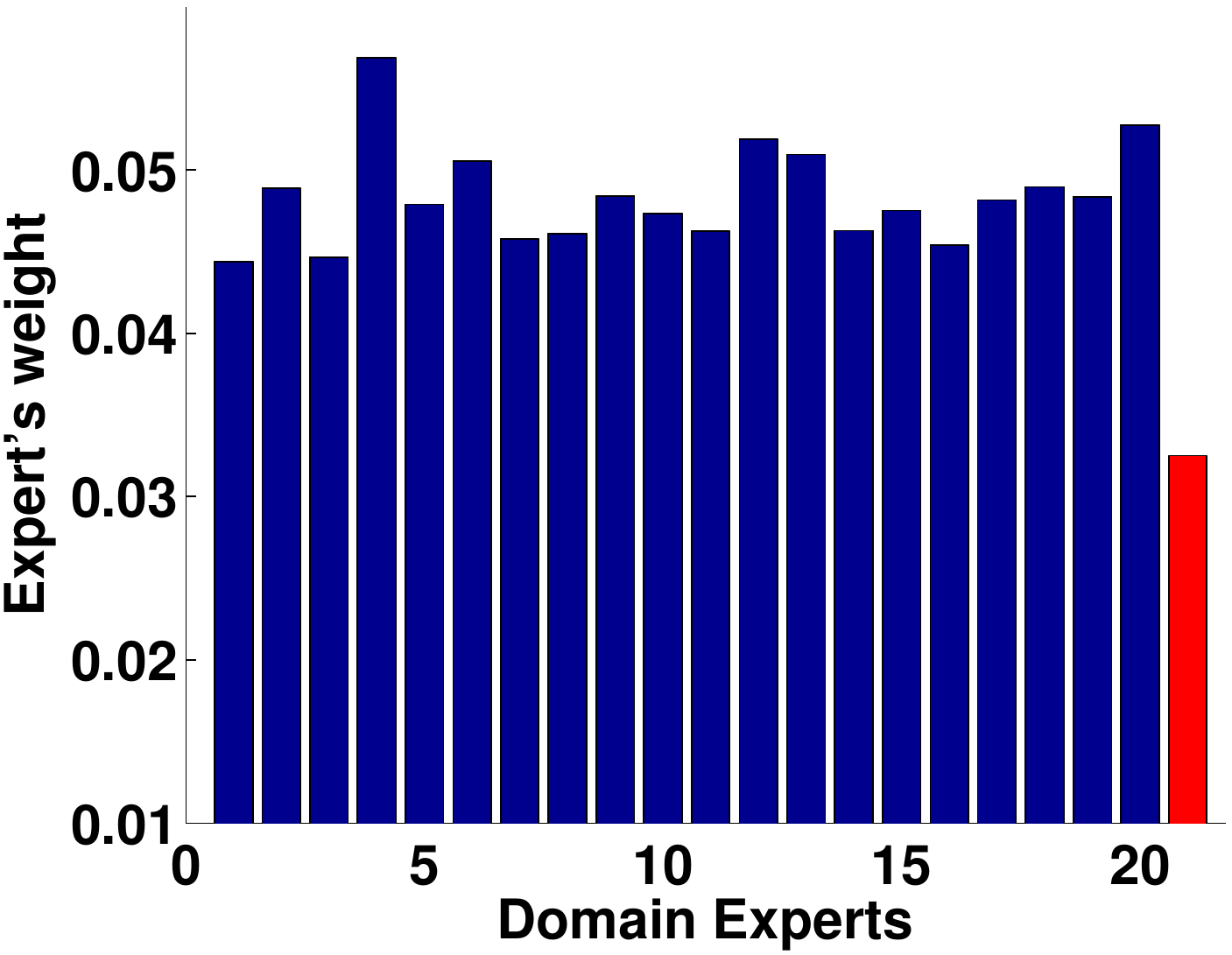}\\
\rot{\rlap{\qquad \quad test subj.\# 22 }} &\includegraphics[scale=.27]{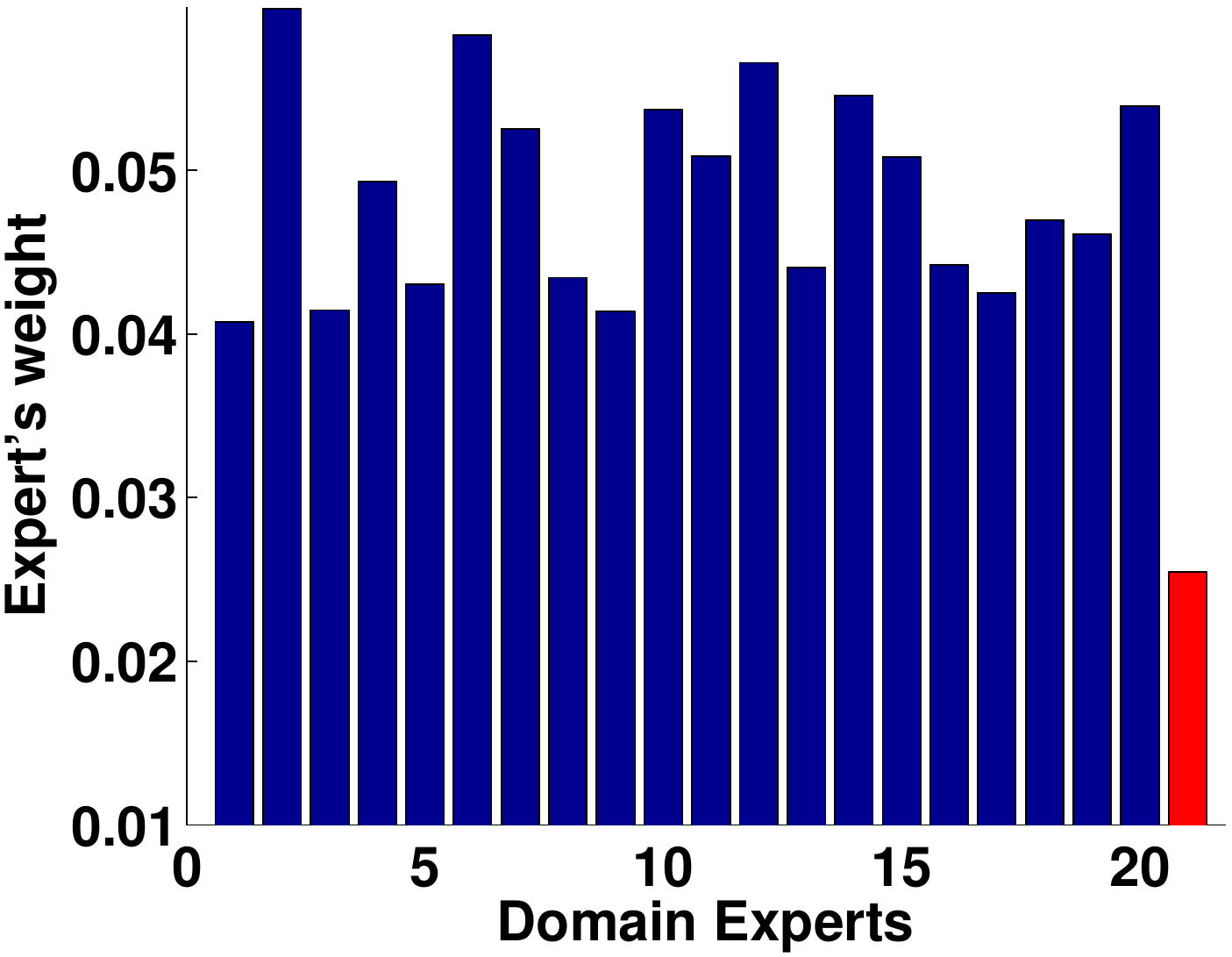} &
\includegraphics[scale=.27]{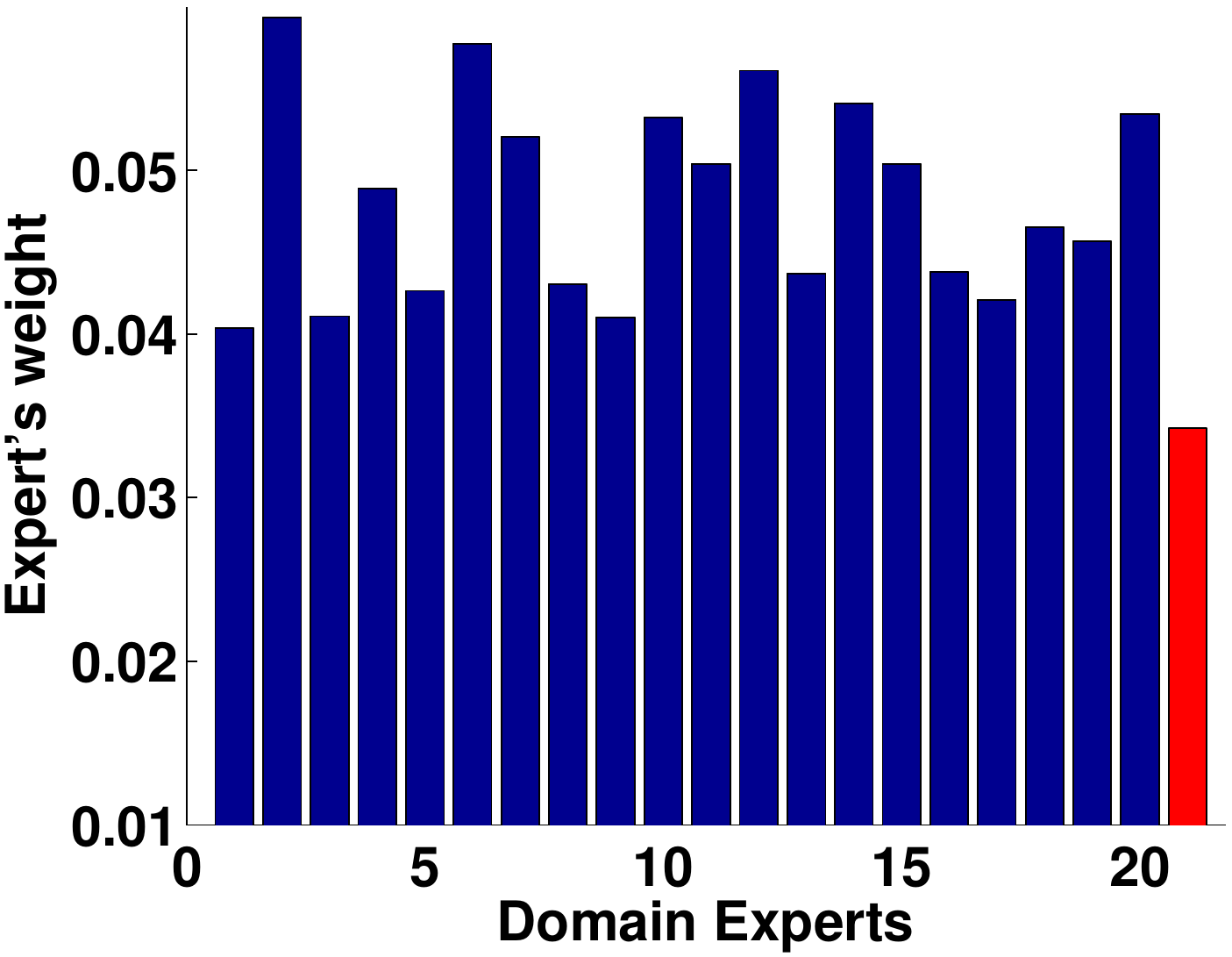}\\
& $N_t = 10$ & $N_t = 50$
\end{tabular}
\caption{{\footnotesize Importance weight of each domain expert (20 source, 1 target) by means of normalized predicted precision for $N_t = 10$ (left) and $N_t = 50$ (right). The confidence of the target specific expert (red/last bar) increases as we increase the cardinality of the labeled target domain data. GPDE correctly predicts the activated AUs, \ie, 12,25, in both cases.}}
\label{fig_w}
\end{figure}

Finally, in Fig.~\ref{fig_w} we demonstrate the ability of the proposed GPDE to fuse the predictions from the individual experts in order to  form the overall prediction. In the selected example, we used the 20 first subjects from DISFA as the source domains, and correctly predicted the ground truth label (AU12 and AU25 active) for 2 different target subjects, \ie, subj.~\#21 and \#22. The depicted weights correspond to the normalized precisions of Eq.~(\ref{post_mugpde}) and indicate a measure of confidence of each domain expert. The importance/confidence of the target expert increases when we use more labeled target data during the adaptation, as expected.
\looseness-1

\section{Conclusions}

The work on domain adaptation in facial behavior analysis is still in its early stage. The conducted experiments on two adaptation tasks (view and subject) indicate several interesting facts: the source classifier trained on a large number of data can easily be outperformed by the classifier trained on as few as 50 examples from the target domain. Furthermore, the existing adaptation approaches try to adapt the target domain to the source domain by assuming that the two distributions can be matched. Yet, as we showed in our experiments on view adaptation, when more target data become available, the target classifier can largely outperform the existing adaptation approaches. The proposed model addresses these challenges by introducing the target expert, allowing it to reach (and outperform) the full performance of either source or target classifiers with as few as 30 target samples. In our future work, we plan to investigate the model adaptation to the other context factors (\ie, `when', `why',`what' and `how'), and also to address modeling of the structure in the output (in the case of AU detection). \looseness-1



{\small
\section*{Acknowledgments}
This work has been funded by the European Community
Horizon 2020 under grant agreement no. 645094 (SEWA), and no.
688835 (DE-ENIGMA).
\looseness-1
\bibliographystyle{ieee}
\bibliography{bibliography}
}

\end{document}